# NL Understanding with a Grammar of Constructions

Wlodek Zadrozny and Marcin Szummer and Stanislaw Jarecki
and David E. Johnson and Leora Morgenstern *
IBM Research Division
T.J.Watson Research Lab
Yorktown Heights, NY 10598 USA

## Abstract

We present an approach to natural language understanding based on a *computable grammar of constructions*. A *construction* consists of a set of features of form and a description of meaning in a context. A grammar is a set of constructions. This kind of grammar is the key element of MINCAL, an implemented natural language speech-enabled interface to an on-line calendar system. The architecture has two key aspects: (a) the use of constructions, integrating descriptions of form, meaning and context into one whole; and (b) the separation of *domain knowledge* (about calendars) from *application knowledge* (about the particular on-line calendar).

## 1 Introduction: an overview of the system

We present an approach to natural language understanding based on a *computable grammar of constructions*. A *construction* consists of a set of features of form and a description of meaning in a context. A grammar is a set of constructions. This kind of grammar is the key element of MINCAL, an implemented natural language speech-enabled interface to an on-line calendar system.

The system consists of a NL grammar, a parser, an on-line calendar, a domain knowledge base (about dates, times and meetings), an application knowledge base (about the calendar), a speech recognizer, a speech generator.

In this paper we describe two key aspects of the system architecture: (a) the use of constructions, where instead of separating NL processing into the phases of syntax, semantics and pragmatics, we integrate descriptions of form, meaning and context into one whole, and use a parser that takes into account all this information (see [10] for details); (b) the separation of the *domain knowledge* (about calendars) and the *application knowledge* (about the particular on-line calendar).

## The dialogs

The system allows users to engage in dialogs like:
— *Schedule a meeting with Bob!*
— *At what time and date?*
— *On August 30th.*
— *At what time?*
— *At 8.*
— *Morning or afternoon?*
— *In the evening.*

The parser recognizes *Schedule a meeting with Bob* as an instance of *sent(imp)*, the imperative construction consisting of a verb and an NP, here *np(event)*. The context is used to prevent another reading in which *with Bob* modifies *schedule*, as in *Dance a tango with Bob!*. That is, a contextual rule is used which says that for calendar applications, people do not modify actions or places. Context also plays an important role in understanding answers, e.g. *At 8*. This is understood as a time expression (and not place or rate or something else) only because of the context.

The parameters of a meeting can be given in many ways, e.g. synonyms or different constructions can be used, users can include as many parameters in a sentence as they wish, and the parameters can be given in any order. As a result there are about 10,000 ways of scheduling meetings (with a given set of parameters).

## How are the dialogs understood

With respect to parsing, grammars of constructions can be parsed like "standard" grammars, except that the set of features is richer. Given a string (representing a sentence, a fragment of a discourse or a paragraph), the parser assigns it a construction. From this viewpoint, the situation is similar to "regular" parsing, and the possible algorithms are similar. We have implemented a prototype chart parser for construction grammars, discussed further in Section 3.

But, clearly, having understood the sentence as a linguistic entity in isolation is not the ultimate goal. Here the message of an utterance must be understood in the context of an intended action. This is done in two steps. First, the system determines the intended action and its parameters, using domain knowledge (meetings+time+places). Second, once all the parameters have been extracted from the dialog, the system executes the action. To do this, the program uses application-specific knowledge to translate the action and its parameters into a form that can be executed by the application (Xdiary).

## 2 Constructions as data structures

A construction is given by the matrix:

$$\begin{bmatrix} \mathbf{N} & : name\_of\_construction \\ \begin{bmatrix} \mathbf{C} & : & context \\ \mathbf{V} & : & structure \\ \mathbf{M} & : & message \end{bmatrix} \end{bmatrix}$$

The *vehicle* **V** consists of formulas describing presence (or perhaps absence) of certain *taxemes*, or features of form, within the structure of the construction. Such a structure is given by a list of subconstructions and the way they have been put together (in all our examples this is concatenation, but there are other possibilities, e.g. wrapping). The *context*, **C**, consists of a set of semantic and pragmatic constraints limiting the application of the construction. It can be viewed as a set of *preconditions* that must be satisfied in order for a construction to be used in parsing. The *message*, **M**, describes the meaning of the construction, via a set of syntactic, semantic and pragmatic constraints.

To make this concrete, let us consider a few examples. We begin with a simple "command construction" consisting of an action verb followed by its argument.

$$\begin{bmatrix} \mathbf{N} & : sent(cmnd, v.np) \\ \begin{bmatrix} \mathbf{C} & : [<hr\ attends> = sr] \\ \mathbf{V} & : \begin{bmatrix} struc = (V.NP) \\ <V\ cons\_n> = verb \\ <V\ \mathbf{M}\ v\_type> = action\_verb \\ <NP\ cons\_n> = np \end{bmatrix} \\ \mathbf{M} & : \begin{bmatrix} sem\_cat = command \\ a\_type = <V\ \mathbf{M}\ sem\_type> \\ a\_obj = <NP\ \mathbf{M}\ sem\_type> \\ agent = hr \end{bmatrix} \end{bmatrix} \end{bmatrix}$$

The context of the construction describes all situations in which the the hearer $hr$ (human or machine) is paying attention to the speaker $sr$ (a "ready" state). The feature *struc* is a list of variables and/or words/tokens; it is used to describe the structure of a construction, and its role is similar to a rule in a generative grammar. (We will write names of variables in capital letters, e.g. $NP$, inside matrices of constructions). The attribute *cons_n* gives the name of a construction that could be assigned to a string. We use it here to say that the form of the construction can be described as a concatenation of two strings, of which one is a verb (construction) and the other an np (construction). Furthermore, the verb type $<V\ \mathbf{M}\ v\_type>$ is "action_verb". (The expression $<V\ \mathbf{M}\ v\_type>$ should be read "the $v\_type$ of the message of $V$").

The message **M** describes the meaning of the construction as that of a command in which the type of action is described by the meaning of the verb, and the object of the action is given by the meaning of the noun phrase. The attribute *sem_type* stands for the "semantic type" and we identify it currently with the word sense. Thus "erase the file" is understood as a command to *delete* the file, if $<erase\ \mathbf{M}\ sem\_type> = delete$, but "erase the picture" might refer to the type of action associated with $rub\_out$. In both cases the hearer $hr$ is supposed to be the agent of the action.

**Constructions: from words to discourse**
Words, phrases, and fragments of discourse can be analyzed as constructions. We view languages as collections of constructions which range from words to discourse. We claim that the same representation scheme can be used for all constructions.

The examples we are going to present have been developed with a specific purpose in mind, namely for scheduling calendar events. In other papers ([10] and [6]), we have presented examples showing that we can give a good descriptions of non-standard constructions. However, in either case descriptions of meanings and contexts are general, and hence applicable to other tasks.

We now turn our attention to words. The verb "cancel" can be represented as follows:

$$\begin{bmatrix} \mathbf{N} & : verb(cancel) \\ \mathbf{C} & : \begin{bmatrix} lang\_code & = & english \\ lang\_channel & = & text \end{bmatrix} \\ \mathbf{V} & : struc = (cancel) \\ \mathbf{M} & : \begin{bmatrix} cat & = & verb \\ sem\_type & = & delete \\ v\_type & = & action\_verb \end{bmatrix} \end{bmatrix}$$

Notice that even simple words require context to be (properly) interpreted. In **C** we say that English text is expected (but in other cases it could also be French text, or French speech, etc.). Some aspects of context do not have to be explicitly specified and can be replaced by defaults.

Although the vehicle and the message are both very simple in this example, the simplicity of the message is a result of deliberate simplification. We have restricted it to the specification of the semantic type, identified with one sense of the word, and to describing the verb type of "cancel" as a verb of action. Notice that the other sense of "cancel" – "offset, balance out" – would appear in another entry.

Of course, in reality, the lexical meaning of any word is a much more complicated matter [1]. For instance, in our lexicon the messages of words may contain many of the attributes that appear in the explanatory combinatorial dictionary of Melcuk [7].

<u>Discourse constructions</u>: To illustrate discourse constructions, we consider the following dialog:

*Have you arranged the room yet?*
*No, but I'll do it right away.*

We view the pattern of the answer *no.but.S* as a discourse construction. It can represented by the following array of features:

$$\left[\begin{array}{l} \mathbf{N} \ : sent(assrt, no.but.S) \\ \left[\begin{array}{l} \mathbf{C} \ : [< p\_utter\ cons\_n > = sent(ques, *)] \\ \mathbf{V} \ : \left[\begin{array}{l} struc = (no.but.S) \\ < S\ cons\_n > = sent(assrt, *) \end{array}\right] \\ \mathbf{M} \ : \left[\begin{array}{l} < p\_sent\ truth\_value > = 0 \\ < S\ \mathbf{M} > \end{array}\right] \end{array}\right] \end{array}\right]$$

As we can see, the construction applies only in the context of a previously asked question, and its message says that the answer to the question is negative, after which it elaborates the answer with a sentence $S$.

## 3  System Architecture

### The parts

MINCAL consists of a NL grammar, a parser, a domain knowledge base (about dates, times and meetings), an on-line calendar (Xdiary), an application knowledge base (about Xdiary), a continuous speech recognizer (IBM, ICSS), a speech generator (Speech Plus, Text to Speech Converter), and the interfaces.

At present, the grammar consists of a few hundred lexical constructions, and about 120 "productions", i.e. constructions describing combinations of other constructions. [1] It covers the basic forms of assertive sentences, but it emphasizes commands. Thus a command can, for example, be given either by v.np (also with "please", or "kindly"), or by an assertive sentence recognized as an indirect speech act ("I'd like to ...", "Leora wants you to ...", etc.). The next large group of constructions covers PPs, with particular emphasis on time and places. Finally, it covers a few discourse constructions, since it is important to deal with sentence fragments in dialogs, e.g. understanding "evening" as "in the evening", when it is an answer to the question "when?".

### The interaction of the modules

**The calendar and the application knowledge base:** Xdiary is an on-line calendar for which we have not written a complete interface, but have focused on the three most important functions: appointment, moving, and canceling appointments. Other functions, such as "to do" lists, window management, listing somebody's appointments, etc., can be dealt with in a similar fashion, and we plan to extend the interface to deal with them. At this point the application knowledge base is very simple. It consists of rules that say how to interpret the data given by the semantic interpreter, for instance the rules for formatting parameters and renaming slots (e.g. event_duration → duration). Such rules are necessary, if the distinction between application and domain knowledge is to be maintained.

**The domain knowledge base:** This has two kinds of facts: (1) background ontology, i.e., is, basic facts about time and places, and (2) linguistic knowledge associated with the domain. The former includes such obvious facts as the number of days in a month, which month follows the other, that offices are places etc. The latter includes facts about how the language is used. For example, the *filters* saying that places do not modify people, so that *I want to meet my manager in the cafeteria* can be unambiguously parsed, with "cafeteria" being a meeting place, and not an attribute of the manager.

**The organization of knowledge:** The issue of the organization of knowledge has been discussed at length in [8] and [9] and the formal model developed there is applicable in the present context. At this point, however, this formal model has only been implemented very crudely. Still the model is worth briefly discussing, because the conceptual distinctions made guide our work and have important practical consequences. The most important thing about it is that we discard the model of background knowledge as a logical theory, and replace it by a model consisting of collection of theories and mechanisms for putting them together depending on circumstances. Thus, the usual, two-part logical structures, consisting of a *metalevel* and an *object level*, are augmented by a third level — a *referential level*. The referential level is a partially ordered collection of theories; it encodes background knowledge in a way resembling a dictionary or an encyclopedia.[2]

### Parser, construction grammar and linguistic knowledge

**Parser:** The parser does not produce (syntactic) structural descriptions of sentences. Instead, it computes meaning representations. For example, it converts adjuncts directly into attributes of place, time, participant etc., once they can be computed, and thus the message of the sentence does not contain any information about how these attributes where expressed or about the attachment of PPs that appear in it. For example, the sentence *I want you to arrange a conference in my office at 5* is analyzed as *sent(assert, svoc)*, an assertive sentence consisting of a subject, a verb, an object and a complement.

---

[1] These are constructions we used in MINCAL. In addition in various experiments we have used a few dozen other constructions, e.g. those covering "open idioms" (see Section 4).

[2] As usual, current situations are described on the object level, and the metalevel is a place for rules that can eliminate some of the models permitted by the object level and the referential level.

The latter and the message of the imperative that is passed to *sent(assert, svoc)* does not contain any structural information about the attachment of the PPs. This message is combined with the messages of the verb and the noun, yielding

```
[   den want(other_agent)]
  [   agent hearer]
  [   mental_agent
      [   [   type person]
          [   den speaker]
[   [   action
        [   [   den arrange]
    [   action_object
        [   type event]
        [   den conference]
        [   number 1]
        [   mods
            [   [   det a]
                [   pp_msg
                    [   [   prep at]
                        [   type time(hour)]
                        [   den
                            [   [   hour
                                    [   5 am_or_pm]
                                [   minute 0]
                    [   [   prep in]
                        [   type place ]
                        [   den office]
                        [   mods
                            [   [   det my]
```

This result of parsing is then interpreted by the domain interpreter to produce:

```
***Slots:
[   [   action_name schedule]
    [   event_name
        [   a conference]
    [   event_time
        [   [   minute 0]
            [   hour
                [   5 am_or_pm]
    [   event_place
        [   my office]
```

Application-specific defaults then produce yet another interpretation where, in addition to filling the slots of Xdiary, [ hour [ 5 am_or_pm] ] is interpreted as [ hour [ 17 ] ].

The parser is a chart parser, working left to right, with no lookahead. The grammar is L-attributed, i.e., has has both synthesized and inherited attributes, but each inherited attribute depends only on inherited attributes of the parent or attributes of the sisters to the left. Hence, although the parser does not have a lookahead step at present, such a step can be added following [2].

# 4 Comparisons with related work

Linguistic arguments for constructions-based grammars has been worked out chiefly by Ch. Fillmore and his colleagues (cf. [3]). Their motivation for advocating such an approach comes from the fact that typical generative theories of grammar cannot deal properly with open idioms illustrated by constructions such as:

> *The more carefully you work, the easier it will get.*
> *Why not fix it yourself?*
> *Much as I like Ronnie, I don't approve of anything he does.*
> *It's time you brushed your teeth.*
> *Him be a doctor?*

The same is true about even so-called robust parsers of English. The reason for this failure can be attributed to the fact that expressions like these "exhibit properties that are not fully predictable from independently known properties of its lexical make-up and its grammatical structure" – [3], p.511. However we do not need a list of "strange" constructions to conclude that thoroughly integrating syntax with semantics and pragmatics could provide us with a better handle on natural language understanding. On a closer examination "normal" constructions exhibit enough complexity to warrant the new approach (see [10] for details).

Jurafsky [4] has independently come up with a proposal for a computable grammar of constructions. We compare our work with his in [10]. Here, we limit ourselves to a few remarks. What is common in both approaches is the centrality of the concept of grammatical construction as a data structure that represents lexical, semantic and syntactic knowledge. However, there are important differences between the two formalisms. First, the actual data structures used to represent constructions are different. The most important difference has to do with the presence of the *context* field in our version of the construction grammar. This allows us to account for the importance of pragmatics in representing many constructions, and to deal with discourse constructions.

Secondly, while Jurafsky acknowledges the need for abstract constructions (pp.43-51), his abstract constructions (*weak constructions*) are not first class citizens — they are defined only extensionally, by specifying the set of constructions they abstract over, and their abstract meaning (e.g. **entity** for NOUN). They are used to simplify descriptions of constituents of other constructions. However, because they do not have a separate *vehicle* part, they cannot be used to assign default meanings. For instance, since *verb* is defined as a collection of all verbs *is + read + cancel + know + look-up + ...*, it cannot be assigned a feature *action_verb* without introducing a contra-

diction – its semantics is therefore given as RELATION/PROCESS. For us the important feature of "abstract" constructions is not that they simplify descriptions of other constructions, but that they have default meanings. (A similar critique of [5] can be found in [10]).

## 5 Summary of results

Our approach to NLU is based both on linguistic arguments and on our dissatisfaction with the state of the art. State of the art systems typically are too "syntax-driven", failing to take context into account in determining the intended meaning of sentences. A related further weakness is that such systems are typically "sentence oriented", rather than "conversation/discourse oriented". In our view, this makes even the most robust systems "brittle" and ultimately impractical.

To test whether a construction-based approach is feasible built a "complete" working system that would include a representation for constructions. To do this, we focused on the "calendar domain", a domain with limited complexity and simple but not uninteresting semantics. We have chosen to deal with simple actions, and not e.g. with question answering, where deeper understanding would be necessary. [3]

**Our contributions:**
1. We have proposed *a new kind of grammar* – computable construction grammars, which are neither semantic, nor syntactic. Instead, their "productions" combine lexical, syntactic, semantic and pragmatic information. [4]
2. We have described *data structures for constructions*, and have shown that they can be effectively used by the parser. Note that the same data structure is used to encode the lexicon and the "syntactic" forms.
3. We have shown *how to parse with constructions*. We have implemented a simple chart parsing algorithm, which can be easily extended to an Eearly-like parser, as long as the construction grammar remains L-attributed. We have found that even a simple parser of construction can be quite efficient. This is partly due to the fact that it does not require copying of all syntactic and semantic information from daughters to mothers; the goal of parsing consists in producing an interpretation, and structural information can be discarded once an interpretation of a phrase is produced. It is also worth emphasizing that invoking domain semantics drastically reduces the number of parses constructed.
4. We have proposed *a modular architecture for NL interfaces* based on the division between linguistic knowledge, domain knowledge base, and application knowledge base. Based on our experience, we believe that this architecture should work in general for speech-enabled interfaces for restricted domains.

---

[3] We have also thought about another possibility, that is, enhancing an IR system, e.g. with the understanding of date expressions.

[4] In what sense are they "computable"? Although this adjective might suggest a formal model with computational complexity results, etc., what we have in mind is pretty trivial: (1) the system actually computes the messages of grammatical construction; (2) the grammars and constructions are well defined data structures, and parsing (combining all associated constructions in all possible ways) is decidable.